\documentclass[twocolumn]{article}
\usepackage{amsmath, amssymb, graphicx, booktabs}
\usepackage{algorithm, algpseudocode}
\usepackage{hyperref}
\usepackage{listings}
\usepackage{xcolor}
\usepackage{caption}
\usepackage{subcaption}
\usepackage{pgfplots}
\pgfplotsset{compat=1.18}
\usepackage{tikz}
\usetikzlibrary{arrows.meta, positioning}
\usepackage{tikz}
\usetikzlibrary{shapes.geometric, arrows.meta, positioning}
\usepackage{multirow}

\newcommand{\method}{Stress-Aware Learning}
\newcommand{\met}{SAL}
\newcommand{\optimizer}{Plastic Deformation Optimizer}
\newcommand{\opt}{PDO}

\definecolor{codegray}{gray}{0.9}
\title{Stress-Aware Resilient Neural Training}

\author{
Ashkan Shakarami\textsuperscript{1}\thanks{Corresponding author: ashkan.shakarami@phd.unipd.it}, 
Yousef Yeganeh\textsuperscript{2},
Azade Farshad\textsuperscript{2},
Lorenzo Nicolè\textsuperscript{1},\\ 
Stefano Ghidoni\textsuperscript{1}, 
Nassir Navab\textsuperscript{2}
}

\date{
\small \textsuperscript{1}University of Padova, Italy\\
\small \textsuperscript{2}Technical University of Munich, Germany\\[1ex]
\small \texttt{
ashkan.shakarami@phd.unipd.it, 
y.yeganeh@tum.de,
azade.farshad@tum.de,
lorenzo.nicole@phd.unipd.it,\\
stefano.ghidoni@unipd.it,
nassir.navab@tum.de
}
}
\begin{document}

\maketitle
\begin{abstract}
\noindent This paper introduces \textbf{\method{}}, a resilient neural training paradigm in which deep neural networks dynamically adjust their optimization behavior—whether under stable training regimes or in settings with uncertain dynamics—based on the concept of \emph{Temporary (Elastic)} and \emph{Permanent (Plastic)} Deformation, inspired by structural fatigue in materials science. To instantiate this concept, we propose \textbf{\optimizer{}}, a stress-aware mechanism that injects adaptive noise into model parameters whenever an internal stress signal—reflecting stagnation in training loss and accuracy—indicates persistent optimization difficulty. This enables the model to escape sharp minima and converge toward flatter, more generalizable regions of the loss landscape. Experiments across six architectures, four optimizers, and seven vision benchmarks demonstrate improved robustness and generalization with minimal computational overhead. The code and 3D visuals will be available on GitHub: \url{https://github.com/Stress-Aware-Learning/SAL}.

\textbf{Keywords:} \method{}, Resilient Neural Training, Optimization, Plastic Deformation. \end{abstract}

\section{Introduction}
\label{sec:introduction}

Deep Neural Networks (DNNs) have achieved transformative success across domains such as computer vision, natural language processing, and speech recognition. Despite their empirical performance, DNN training remains sensitive to optimization dynamics and prone to stagnation, convergence to sharp minima, or overfitting. Standard optimizers such as SGD~\cite{sgd_momentum}, Adam~\cite{adam}, RMSProp~\cite{rmsprop}, and Nadam~\cite{dozat2016incorporating} operate based on fixed update rules that respond only to local gradient statistics. They lack mechanisms to dynamically adjust optimization strategies based on broader training feedback, such as stagnation in performance or structural learning difficulty.

Yet, the loss landscape in DNNs is inherently non-stationary: gradients fluctuate, curvature evolves, and performance plateaus frequently occur. In the absence of internal mechanisms for adaptation, models risk settling into brittle solutions. Various methods attempt to mitigate these issues. Regularization techniques such as Dropout~\cite{dropout}, Stochastic Depth~\cite{stochastic_depth}, and label smoothing~\cite{szegedy2016rethinking} inject stochasticity to prevent co-adaptation. Techniques like Sharpness-Aware Minimization (SAM)~\cite{foret2020sharpness} and Stochastic Weight Averaging (SWA)~\cite{izmailov2018averaging} promote flatter minima, but remain static and do not take advantage of internal feedback from the training process.

Stochastic optimization strategies such as Entropy-SGD~\cite{entropy_sgd}, SGLD~\cite{sgld}, and more recent adaptive noise methods~\cite{xie2020noise, liu2021training} introduce perturbations to escape sharp minima. However, these approaches typically rely on externally defined heuristics or fixed schedules. Meta-learning frameworks, including hypergradient-based optimization~\cite{baydin2018online} and MAML~\cite{finn2017model}, offer more flexible training dynamics, but often involve complex inner-loop objectives and lack persistent indicators of optimization stagnation. Approaches aimed at robustness—such as adversarial training~\cite{madry2018towards}, distributional robustness~\cite{sinha2018certifiable}, and Jacobian regularization~\cite{jakubovitz2018improving}—depend on predefined robustness objectives, which may not generalize across tasks.

To overcome these limitations, we introduce \textbf{\method{} (\met{})} (\autoref{sec:Stress-Aware Learning}), a training framework that enables self-regulation based on an internally accumulated \emph{stress signal}. This signal quantifies the stagnation of optimization over time and modulates the intensity of interventions during training. When the stress signal exceeds specific thresholds, \met{} applies proportionally scaled perturbations—encouraging exploration in flat regions of the loss landscape and facilitating escape from sharp, brittle solutions. This mechanism is inspired by material fatigue dynamics in structural mechanics, drawing an analogy between training difficulty and cumulative plastic deformation under stress \cite{deformation_nie2021review, deformation_kim2024effect,deformation_li2024modeling,deformation_tang2024improved}.

At the core of \met{} lies \textbf{\optimizer{} (\opt{})} (\autoref{sec:pdo}), a lightweight and differentiable regularization strategy that adaptively injects noise or applies structural updates to model parameters based on an internal stress signal. In contrast to global optimization heuristics such as Particle Swarm Optimization~\cite{kennedy1995particle}, Genetic Algorithms~\cite{holland1975adaptation}, Grey Wolf Optimizer~\cite{mirjalili2014grey}, and Simulated Annealing~\cite{kirkpatrick1983optimization}—which are typically non-differentiable, computationally costly, and lack introspective feedback—\opt{} integrates seamlessly with modern gradient-based optimizers like Adam and maintains end-to-end differentiability (\autoref{sec:optimizer_adaptability}).

A motivation behind our work lies in addressing the wide spectrum of conditions under which deep networks are trained. In practical settings, training environments range from carefully tuned, stable setups—featuring deep architectures, low learning rates, and high-quality datasets—to challenging, unstable ones, where data may be scarce, hyperparameters poorly selected, and architectures shallow. Standard optimizers and regularization methods often fail to adapt under such adverse configurations, leading to optimization collapse or sharp minima convergence. \met{} is designed to operate effectively across this continuum. Its internal stress-aware feedback mechanism offers robustness not only in well-behaved training regimes but also in suboptimal or unpredictable ones (\autoref{sec:Robustness in Stable vs. Uncertain Training Conditions}). In fact, its adaptive interventions become especially pronounced in unstable conditions, where conventional training often stagnates (\autoref{fig:comparison_unstable}). This makes \met{} a practical and generalizable tool when training dynamics are uncertain or difficult to manually tune.

\section{\met{}}
\label{sec:Stress-Aware Learning}
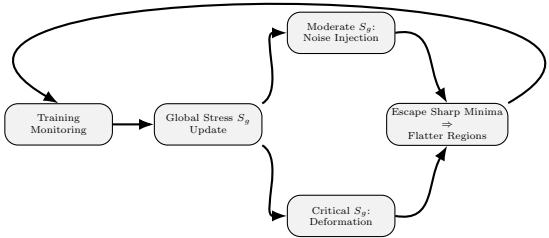
\begin{figure}[h]
\noindent
\begin{minipage}{\columnwidth}
\centering
\makebox[\textwidth][c]{ 
\begin{tikzpicture}[
    scale=0.55, transform shape,
    every node/.style={align=center},
    box/.style={
        rectangle, draw, rounded corners=5pt,
        minimum width=2.6cm, minimum height=1.0cm,
        fill=gray!10, font=\scriptsize
    },
    arrow/.style={-{Latex[length=2mm]}, thick}
]

\node[box] (mon) {Training\\ Monitoring};
\node[box, right=of mon] (stress) {Global Stress $S_g$\\ Update};
\node[box, above right=of stress, xshift=-0.4cm, yshift=0.2cm] (mild) {Moderate $S_g$:\\ Noise Injection};
\node[box, below right=of stress, xshift=-0.4cm, yshift=-0.2cm] (crit) {Critical $S_g$:\\ Deformation};
\node[box, right=3.0cm of stress] (esc) {Escape Sharp Minima \\ $\Rightarrow$ \\Flatter Regions};

\draw[arrow] (mon.east) -- (stress.west);
\draw[arrow, out=30, in=180, looseness=0.9] (stress.north east) to (mild.west);
\draw[arrow, out=-30, in=180, looseness=0.9] (stress.south east) to (crit.west);

\draw[arrow, out=10, in=120, looseness=1.2] (mild.east) to (esc.north);
\draw[arrow, out=-10, in=240, looseness=1.2] (crit.east) to (esc.south);

\draw[arrow, out=30, in=150, looseness=1.5] (esc.north east) to (mon.north);

\end{tikzpicture}
}
\caption{\met{} scheme. Stress accumulates from Training Monitoring and triggers Noise Injection or Deformation (plastic deformation) responses, and the resulting escape to flatter minima feeds back into renewed monitoring.}
\label{fig:sal_overview}
\end{minipage}
\end{figure}

Shown in \autoref{fig:sal_overview}, \met{} introduces a closed-loop optimization scheme inspired by material fatigue \cite{deformation_nie2021review,deformation_kim2024effect}, where learning dynamics are continuously shaped by an internally maintained scalar signal, the global stress $S_g$ (\autoref{sec:Global Stress Accumulation}). This signal evolves based on epoch-wise improvements in loss and accuracy, serving as a proxy for training difficulty. The mechanism operates in two distinct phases depending on the current stress level:

\textbf{Moderate Stress ($S_g < S_{\text{yield}}$):} The model applies small stochastic perturbations to its parameters to encourage exploration and escape sharp minima from early stages.

\textbf{Critical Stress ($S_g \geq S_{\text{yield}}$):} The model undergoes a more substantial transformation—termed plastic deformation—to redirect convergence away from persistent suboptimal regions. This simulates an irreversible shift in parameter space, akin to structural yielding in physical systems.

These interventions are neither manually scheduled nor externally triggered. Instead, \met{} dynamically adapts its regularization intensity based on the optimization trajectory (\autoref{fig:3d_trajectory} in \autoref{sec:appendix} shows a visual example of this), ensuring self-regulated behavior throughout training. After each intervention, the system resets the stress and resumes monitoring, thus completing a feedback-driven cycle that aligns the optimization effort with the learning progress.

\paragraph{Global Stress Accumulation.} 
\label{sec:Global Stress Accumulation}
At the heart of \met{} lies the global stress scalar $S_g \in [0, S_{\text{max}}]$, which evolves based on the observed training dynamics. Specifically, $S_g$ jointly monitors the epoch-to-epoch improvements in loss ($\ell_e$) and accuracy ($\text{Acc}_e$) according to the following update rule:

\begin{equation}
S_g \leftarrow 
\begin{cases}
\max(0, S_g - \rho), & 
\begin{aligned}[t]
&\text{if } \ell_{e-1} - \ell_e > \epsilon_{\text{loss}} 
\text{ and}\\
&\quad \text{Acc}_e - \text{Acc}_{e-1} > \epsilon_{\text{acc}}
\end{aligned} \\
\min(S_{\text{max}}, S_g + \theta), & \text{otherwise}
\end{cases}
\label{eq:stress_update}
\end{equation}

Here, $\rho$ and $\theta$ control the decay and accumulation rates of stress, while $\epsilon_{\text{loss}}$ and $\epsilon_{\text{acc}}$ define minimal thresholds for considering a step successful. Thus, stress naturally decays when substantial progress is made and accumulates when training stagnates, enabling the system to adaptively escalate interventions based on endogenous optimization signals.

\subsection{\opt{}}
\label{sec:pdo}
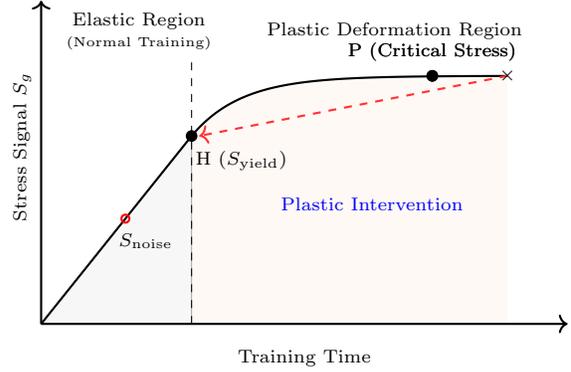
\begin{figure}[ht]
\centering
\begin{tikzpicture}[
    every node/.style={align=center},
    elasticregion/.style={fill=gray!15, opacity=0.4},
    plasticregion/.style={fill=orange!10, opacity=0.4},
    lab/.style={font=\scriptsize},
    point/.style={fill=white, circle, draw, inner sep=1pt}
]

\draw[elasticregion] (0,0) -- (2,2.5) -- (2,0) -- cycle;

\begin{scope}
  \clip (2,0) -- (2,2.5)
        -- plot[domain=2:6.2, smooth, variable=\x, samples=100]
        ({\x}, {2.5 + 0.8*(1-exp(-1.5*(\x-2)))})
        -- (6.2,0) -- cycle;
  \fill[plasticregion] (2,0) rectangle (6.2,4);
\end{scope}

\draw[->, thick] (0,0) -- (7,0) node[midway, below=6pt] {\scriptsize Training Time};
\draw[->, thick] (0,0) -- (0,4.3) node[midway, rotate=90, above=6pt] {\scriptsize Stress Signal $S_g$};

\draw[dashed, thin] (2,0) -- (2,3.5);

\draw[thick] (0,0) -- (2,2.5); 
\draw[thick, domain=2:6.2, smooth, variable=\x, samples=100]
    plot ({\x}, {2.5 + 0.8*(1-exp(-1.5*(\x-2)))});
\draw[->, thick, red!80, dashed] (6.2,3.3) -- (2.1,2.5);
\draw[red, thick] (5.2,3.3) circle (1.5pt);
\filldraw[black] (5.2,3.3) circle (2pt) node[above=2pt, lab] {P (Critical Stress)};

\draw[red, thick] (1.12,1.4) circle (1.5pt);
\node[below left=2pt and -2pt, lab] at (1.8,1.4) {$S_{\text{noise}}$};

\filldraw[black] (2,2.5) circle (2pt) node[below right=2pt and -2pt, lab] {H ($S_{\text{yield}}$)};
\filldraw[black] (5.2,3.3) circle (2pt) node[above=2pt, lab] {P (Critical Stress)};

\node[lab] at (1.3,3.9) {Elastic Region\\\tiny (Normal Training)};
\node[lab] at (4.7,3.9) {Plastic Deformation Region};

\node[lab, text=blue] at (4.4,1.6) {Plastic Intervention};
\node[lab, text=black, font=\bfseries] at (6.2,3.3) {\scriptsize $\times$};

\end{tikzpicture}
\caption{\opt{}. Stress progression in \met{}. When the internal stress signal $S_g$ exceeds a soft threshold $S_{\text{noise}}$, Gaussian noise is injected to encourage exploration. If $S_g$ surpasses the yield point $S_{\text{yield}}$, the system enters the plastic regime. In cases where plastic intervention fails to improve training, the model resets to the yield point (shown by the red arrow on the figure), enabling recovery.}

\label{fig:sal_pdo_final}
\end{figure}

At the core of \met{} lies the \opt{}, which transforms the accumulated stress signal $S_g$ into real-time perturbations that regulate training behavior. As illustrated in ~\autoref{fig:sal_pdo_final}, \opt{} operates in two progressive regimes, adapting to optimization difficulty based on the internal state of the model.

\paragraph{Moderate Stress Phase.}
When $S_g$ surpasses a soft threshold $S_{\text{noise}}$—indicative of early stagnation—\opt{} injects Gaussian noise to encourage exploration and escape from sharp local minima:
\begin{equation}
w \leftarrow w + \alpha (\Delta + \lambda S_g) \cdot \mathcal{N}(0, 1), \quad \alpha = \min\left(1, \frac{S_g}{S_{\text{yield}}}\right)
\label{eq:perturbation}
\end{equation}
Here, $\Delta$ defines a base noise level, $\lambda$ adjusts the magnitude of noise to the stress level, and $\alpha$ ensures that perturbations scale smoothly as $S_g$ approaches the yield threshold $S_{\text{yield}}$.

\paragraph{Critical Stress Phase.}
If training stagnation persists and $S_g$ exceeds $S_{\text{yield}}$, the system enters the plastic deformation regime. \opt{} applies a more invasive intervention to the final-layer weights:
\begin{equation}
w_{\text{final}} \leftarrow 0.9 \cdot w_{\text{final}} + \mathcal{N}(0, 0.02)
\label{eq:yield}
\end{equation}
This simulates a structural shift in parameter space, effectively re-routing the convergence path. After deformation, the stress signal is reset to zero ($S_g \leftarrow 0$), closing the feedback loop.

\paragraph{Failure Recovery.}
As shown by the reset arrow in ~\autoref{fig:sal_pdo_final}, if the plastic intervention fails to improve training dynamics, the model reverts to the yield point. This mechanism prevents prolonged divergence and enables recovery into the elastic regime.

\opt{} enables \met{} to operate as an introspective, self-correcting training system. Unlike heuristic-based strategies \cite{kennedy1995particle,holland1975adaptation,mirjalili2014grey,kirkpatrick1983optimization}, it leverages internal training feedback for real-time modulation of noise and structural adjustment, without modifying the base optimizer.

\subsection{Integrated Training Procedure}
\label{sec:training}
The \met{} procedure is incorporated directly into the standard training loop, augmented by the \opt{}. This integration enables model parameters to undergo targeted interventions when training progress becomes insufficient, as quantified by an internal scalar stress signal \( S_g \).

At each training epoch, the model proceeds with conventional weight updates using Adam. Following this, average epoch-level metrics—specifically the loss \( \ell_e \) and accuracy \( \text{Acc}_e \)—are computed to assess training progress. These metrics are then used to update \( S_g \) via ~\autoref{eq:stress_update}, which increases stress when performance stagnates and decays it upon sufficient improvement.

If the stress level exceeds the moderate noise threshold \( S_{\text{noise}} \), controlled perturbations are applied to model parameters (\autoref{eq:perturbation}). If the stress surpasses the yield threshold \( S_{\text{yield}} \), a stronger plastic adaptation is triggered according to ~\autoref{eq:yield}, and \( S_g \) is reset.

The full training loop, including stress accumulation and conditional interventions, is detailed in Algorithm ~\ref{alg:sal}. This procedure does not modify the base optimizer but augments it with adaptive mechanisms that respond directly to observed training behavior.

\begin{algorithm}[h]
\caption{\met{} via \opt{}}
\label{alg:sal}
\begin{algorithmic}[1]
\State Initialize \( S_g = 0 \), \( \ell_{\text{prev}} = \infty \), \( \text{Acc}_{\text{prev}} = 0 \)
\For{epoch \( e = 1 \) to \( E \)}
    \For{mini-batch \( (x, y) \)}
        \State Compute loss \( \ell \), update weights using Adam
    \EndFor
    \State Compute average epoch loss \( \ell_e \), accuracy \( \text{Acc}_e \)
    \State Update stress \( S_g \) using ~\autoref{eq:stress_update}
    \If{ \( S_g > S_{\text{noise}} \) }
        \State Apply parameter perturbations via ~\autoref{eq:perturbation}
    \EndIf
    \If{ \( S_g > S_{\text{yield}} \) }
        \State Apply plastic yield via ~\autoref{eq:yield}; set \( S_g \leftarrow 0 \)
    \EndIf
    \State Update history: \( \ell_{\text{prev}} \leftarrow \ell_e \), \( \text{Acc}_{\text{prev}} \leftarrow \text{Acc}_e \)
\EndFor
\end{algorithmic}
\end{algorithm}

\subsection{Theoretical Justification}
\label{sec:theory-sal}

We formally analyze \met{} by characterizing how stress-regulated perturbations influence convergence. Let \( \mathcal{L}(\mathbf{w}) \) denote the empirical loss over model parameters \( \mathbf{w} \in \mathbb{R}^d \), defined as:

\begin{equation}
\mathcal{L}(\mathbf{w}) := \frac{1}{|\mathcal{D}|} \sum_{(x_i, y_i) \in \mathcal{D}} \ell(f(x_i; \mathbf{w}), y_i)
\end{equation}

where \( f(x_i; \mathbf{w}) \) is the model prediction, \( \ell(\cdot, \cdot) \) is the task loss (e.g., cross-entropy), and \( \mathcal{D} \) is the training dataset. Let \( \mathbf{H}(\mathbf{w}) = \nabla^2 \mathcal{L}(\mathbf{w}) \) denote the Hessian matrix.

\paragraph{Global Stress Signal.} \met{} maintains a scalar variable \( S_g \in [0, S_{\max}] \) representing the accumulated training difficulty. It is updated at each epoch \( t \) based on improvements in loss and accuracy:

\begin{equation}
S_g^{(t+1)} =
\begin{cases}
\max(0, S_g^{(t)} - \rho), & \text{if } \Delta\mathcal{L}^{(t)} > \epsilon_\ell \text{ and}\\
&\quad \Delta\text{Acc}^{(t)} > \epsilon_{\text{acc}} \\
\min(S_{\max}, S_g^{(t)} + \theta), & \text{otherwise}
\end{cases}
\end{equation}

Here, \( \rho, \theta > 0 \) are decay and accumulation rates, and \( \epsilon_\ell, \epsilon_{\text{acc}} \) are thresholds for meaningful improvement in loss and accuracy, respectively.

\subsubsection{Effect of Stress-Modulated Perturbations}

When \( S_g > S_{\text{noise}} \), \opt{} perturbs parameters as:

\begin{equation}
\mathbf{w} \leftarrow \mathbf{w} + \alpha (\Delta + \lambda S_g) \cdot \boldsymbol{\xi}, \quad \boldsymbol{\xi} \sim \mathcal{N}(0, \mathbf{I})
\end{equation}

where \( \Delta \) is a base noise level, \( \lambda \) controls stress sensitivity, and \( \alpha = \min(1, S_g/S_{\text{yield}}) \) scales perturbation relative to yield threshold \( S_{\text{yield}} \).

Using a second-order Taylor expansion:

\begin{equation}
\mathcal{L}(\mathbf{w} + \delta) \approx \mathcal{L}(\mathbf{w}) + \nabla \mathcal{L}(\mathbf{w})^\top \delta + \frac{1}{2} \delta^\top \mathbf{H} \delta
\end{equation}

and assuming \( \delta \sim \mathcal{N}(0, \sigma^2 \mathbf{I}) \), the expected loss becomes:

\begin{equation}
\mathbb{E}[\mathcal{L}(\mathbf{w} + \delta)] \approx \mathcal{L}(\mathbf{w}) + \frac{\sigma^2}{2} \operatorname{Tr}(\mathbf{H})
\label{eq:expected_loss}
\end{equation}

where \( \sigma^2 = \alpha^2 (\Delta + \lambda S_g)^2 \). This shows that higher \( S_g \) increases \( \sigma^2 \), amplifying curvature penalization and encouraging convergence toward flatter regions (lower \( \operatorname{Tr}(\mathbf{H}) \)).

\subsubsection{Plastic Deformation as Distributional Shift}

If \( S_g > S_{\text{yield}} \), the system applies a plastic transformation to final-layer weights:

\begin{equation}
\mathbf{w}_{\text{final}} \leftarrow 0.9 \cdot \mathbf{w}_{\text{final}} + \mathcal{N}(0, \sigma_y^2)
\end{equation}

where \( \sigma_y^2 \) controls the magnitude of injected noise. This simulates structural deformation to escape sharp local minima.

\subsubsection{Convergence Toward Flat Minima}

Assume local curvature is bounded: \( \|\mathbf{H}(\mathbf{w})\|_2 \leq L \). Then, under Gaussian perturbations \( \delta \sim \mathcal{N}(0, \sigma^2 \mathbf{I}) \), it holds:

\begin{equation}
\|\nabla \mathbb{E}[\mathcal{L}(\mathbf{w} + \delta)] - \nabla \mathcal{L}(\mathbf{w})\| \leq \frac{\sigma^2 L}{2}
\end{equation}

Hence, the perturbed gradient biases optimization away from sharp curvature regions. If stress remains active for a non-trivial portion of training (\( \mathbb{E}[S_g^{(t)}] > 0 \)), \met{} satisfies:

\begin{equation}
\lim_{t \to \infty} \mathbb{E}[\operatorname{Tr}(\mathbf{H}(\mathbf{w}^{(t)}_{\text{\met{}}}))] < \mathbb{E}[\operatorname{Tr}(\mathbf{H}(\mathbf{w}^{(t)}_{\text{Adam}}))]
\end{equation}

This establishes that \met{} converges in expectation to flatter minima than conventional training.

\section{Experiments}
\subsection{Implementation Details}
All experiments were conducted using TensorFlow with fixed random seeds to ensure reproducibility. The primary dataset used was Imagenette, resized to \( 64 \times 64 \) resolution. Imagenette is a subset of 10 classes from Imagenet (Tench, English springer, Cassette player, Chain saw, Church, French horn, Garbage truck, Gas pump, Golf ball, Parachute). All inputs were normalized to the \([0,1]\) range. The batch size was set to 64, and models were trained for 50 epochs. Unless otherwise specified, Adam Optimizer with a fixed learning rate  \( 1 \times 10^{-5} \) was used in all settings. Both the baseline and \met{} model shared identical architectures. Evaluation metrics included Top-1 and Top-5 accuracy, test loss, training time, and memory usage. Post-training sharpness was also measured to assess generalization properties. 
We used a DenseNet201 model pre-trained on ImageNet as the backbone. A global average pooling layer followed the feature extractor, then two fully connected layers with 512 and 128 units, respectively, separated by a dropout layer with a dropout rate of 0.5. A final softmax classification layer produced the output probabilities. The stress signal is updated at each epoch:

\begin{align}
S_g &\leftarrow \max(0, S_g - \rho) \quad \text{if improvement,} \notag \\
S_g &\leftarrow \min(1, S_g + \theta) \quad \text{otherwise}
\end{align}

with decay and growth rates set to \( \rho = 0.0005 \) and \( \theta = 0.005 \), respectively.

Interventions are triggered based on two thresholds:
\begin{itemize}
    \item \textbf{Noise Injection:} If \( S_g > S_{\text{noise}} = 0.005 \), Gaussian perturbations are applied across all trainable weights:
    \[
    w \leftarrow w + \alpha (\Delta + \lambda S_g) \cdot \mathcal{N}(0, 1)
    \]
    where \( \alpha = \min(1, \frac{S_g}{S_{\text{yield}}}) \), \( \Delta = 10^{-7} \), and \( \lambda = 10^{-5} \).
    
    \item \textbf{Plastic Deformation:} If \( S_g \geq S_{\text{yield}} = 0.01 \), a structural intervention modifies the last three trainable layers:
    \[
    w \leftarrow 0.9 \cdot w + \mathcal{N}(0, 0.02)
    \]
    followed by a reset \( S_g \leftarrow 0 \).
\end{itemize}

These mechanisms were activated after a warm-up phase of 15 epochs, ensuring initial training stability. SAL’s interventions were logged and aligned with detected stagnation, illustrating its real-time adaptation to learning dynamics (\autoref{fig:sharpness_evolution_plot}).

\subsection{Stable vs. Uncertain Training}
\label{sec:Robustness in Stable vs. Uncertain Training Conditions}
\met{} exhibits robust adaptability across both well-behaved and degraded training regimes through dynamic, stress-aware interventions. Using an internal stress signal, it detects stagnation phases (\autoref{fig:sharpness_evolution_plot}) and responds with corrective actions—such as noise injection and plastic deformation—to maintain optimization momentum and generalization.

\begin{figure}[h!]
  \centering
  \includegraphics[width=0.99\linewidth]{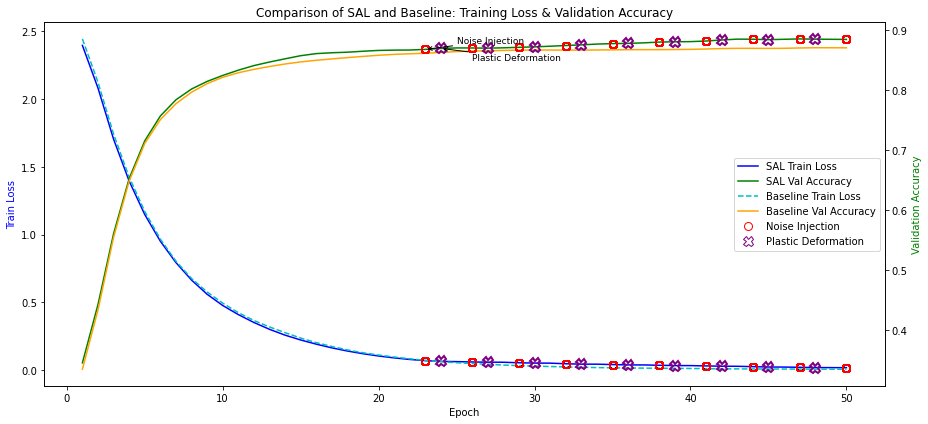}
  \caption{Training dynamics in stable conditions. SAL stabilizes convergence through timely, stress-guided interventions.}
  \label{fig:comparison_stable}
\end{figure}

\paragraph{Stable Training Conditions.} \label{sec:Stable Training Conditions} In a controlled training setup using DenseNet201, the Adam optimizer, a conservative learning rate ($1\mathrm{e}{-5}$), and Imagenette (64$\times$64 input), \met{} demonstrates higher accuracy than the baseline. As shown in~\autoref{fig:comparison_stable}, stress-triggered interventions occur, injecting noise or applying plastic deformation. These interventions are not disruptive but act as minimal, context-aware adjustments. Consequently, validation accuracy improves consistently throughout training, confirming that \met{} enhances resilience without over-regularization.

\begin{figure}[h!]
  \centering
  \includegraphics[width=0.99\linewidth]{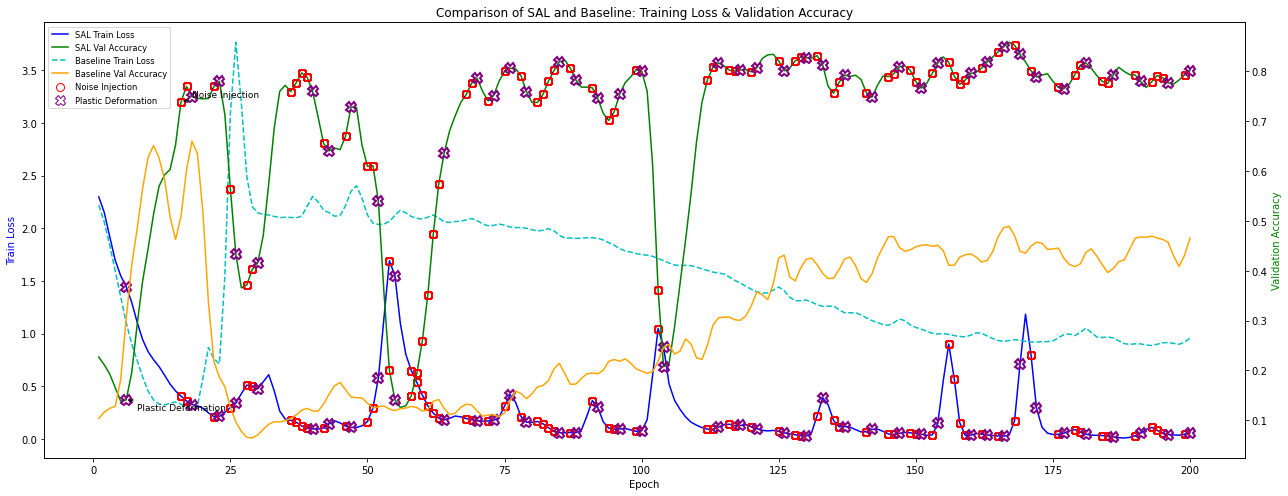}
  \caption{Training dynamics in degraded conditions. SAL rescues performance by activating corrective responses to stagnation.}
  \label{fig:comparison_unstable}
\end{figure}

\paragraph{Uncertain or Degraded Conditions.} In contrast, we evaluate a challenging scenario using ResNet50V2 trained on Corel-1k (32$\times$32 resolution, 80\% training and 20\% validation) with a high learning rate ($1\mathrm{e}{-3}$) and 200 epochs. The baseline optimizer suffers from erratic fluctuations and performance collapse. \met{}, however, maintains coherent learning by activating stress-triggered corrections during instability, injecting noise, or applying plastic deformation to restore gradient flow. As depicted in~\autoref{fig:comparison_unstable}, perturbations occur in tandem with divergence, recovering learning momentum, and achieving higher validation accuracy under unstable dynamics. Learning momentum here refers to the model's ability to sustain progress in optimization despite instability, avoiding stalls or collapse.

These findings highlight that \met{} is effective in well-calibrated environments and remains indispensable in ill-conditioned setups—making it particularly suitable for deployment in scenarios where data quality or training configurations are suboptimal or unknown \cite{shakarami2025depvit, shakarami2025utsp, shakarami2025transformative, shakarami2025velu, shakarami2024aicancer, shakarami2020alzheimer, shakarami2023tcnn,shakarami2021yolo}.

\subsection{Stress Signal Regulation}

\begin{figure}[h!]
  \centering
  \includegraphics[width=0.75\linewidth]{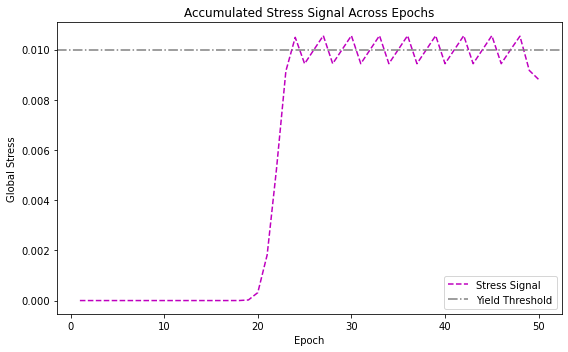}
  \caption{Evolution of accumulated stress signal across training epochs. Stress rises during periods of optimization stagnation and resets following plastic deformation, indicating effective self-regulation by \met{}.}
  \label{fig:accumulated_stress}
\end{figure}

\autoref{fig:accumulated_stress} illustrates the dynamic evolution of the accumulated stress signal \( S_g \) throughout training. The signal exhibits a non-monotonic pattern—gradually increasing during phases of stagnation in training performance and dropping sharply following corrective interventions. This temporal behavior highlights the self-regulatory nature of \met{}, where the stress signal serves as an internal proxy for training difficulty.

Notably, plateaus in loss and accuracy align with peaks in \( S_g \), confirming that the system effectively detects stagnation. The subsequent reduction in stress after plastic deformation or noise injection signifies successful recovery. This cyclical dynamic reflects a key strength of \met{}: it does not rely on fixed schedules or arbitrary heuristics, but instead leverages a principled feedback loop to guide interventions.

The core mechanism lies in the continuous monitoring of optimization progress. If improvements in loss and accuracy stall, stress accumulates; if training resumes successfully, stress decays. Once the stress exceeds predefined thresholds, \met{} triggers proportional responses—ranging from mild noise perturbations to more impactful plastic updates—based on the severity of the stagnation. By anchoring these decisions in real-time training feedback, \met{} ensures that its interventions are both timely and adaptive, promoting stability without compromising learning efficiency.

\subsection{Correlation Between Stress and Loss Dynamics}
\begin{figure}[h!]
  \centering
  \includegraphics[width=0.75\linewidth]{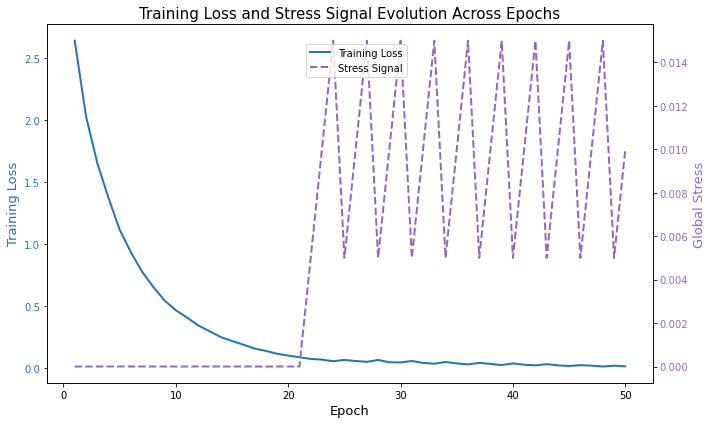}
  \caption{Joint evolution of training loss and global stress signal $S_g$ across epochs. Stress peaks align with loss stagnation, validating the role of stress as an internal signal for targeted optimization interventions.}
  \label{fig:loss_stress_evolution}
\end{figure}

~\autoref{fig:loss_stress_evolution} overlays the trajectory of training loss with the corresponding evolution of the accumulated stress signal $S_g$. A clear temporal correlation emerges: stress consistently increases during epochs where loss plateaus, indicating ineffective optimization progress. These stress spikes serve as internal indicators of stagnation, prompting \met{} to trigger perturbative responses such as noise injection or plastic deformation. When loss reduction resumes following these interventions, the stress signal naturally subsides. This feedback loop enables \met{} to continuously modulate its regularization intensity based on empirical difficulty, rather than relying on predefined schedules.

The alignment between stress dynamics and loss trajectory validates the design rationale of \met{}. The stress signal functions as both a monitoring mechanism and a principled trigger for intervention, enhancing the optimizer's ability to escape flat regions or sharp minima without introducing unnecessary disruption.

\subsection{Sharpness Regulation via Plastic Events}
\begin{figure}[h!]
  \centering
  \includegraphics[width=0.75\linewidth]{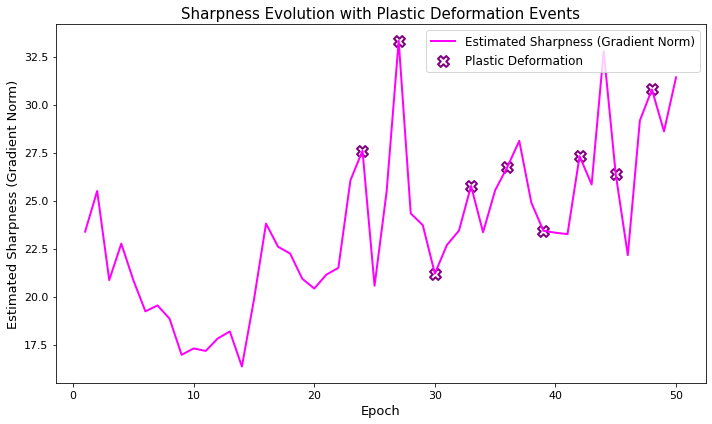}
  \caption{Sharpness progression during training. Plastic deformation events align with sharpness spikes, demonstrating \met{}'s ability to regulate curvature and maintain optimization stability.}
  \label{fig:sharpness_evolution_plot}
\end{figure}

To assess how \met{} modulates training sharpness,~\autoref{fig:sharpness_evolution_plot} presents the estimated sharpness (measured as gradient norm) across epochs, alongside plastic deformation events. Peaks in sharpness correlate closely with the occurrence of plastic interventions, confirming that the internal stress signal effectively identifies unstable or overly sharp regions in the loss landscape. These interventions induce noticeable reductions in sharpness, improving trajectory stability. 

The plot illustrates that \met{} acts as a curvature-aware optimizer, maintaining a balanced path between exploration and convergence. Instead of converging prematurely into narrow and high-curvature minima, the model periodically softens its optimization trajectory through plastic events, which serve as localized smoothing operations. This results in a dynamic regulation mechanism where sharpness is allowed to rise temporarily—encouraging learning—but is promptly suppressed when exceeding a tolerable threshold. Such behavior aligns with the theoretical goals of generalization, as flatter regions are empirically linked to better robustness and performance. Taken together, the sharpness dynamics validate \met{}'s role as a geometry-aware training paradigm that influences both the optimization process and its final convergence behavior.

\subsection{3D Loss Landscapes}
\label{sec:3D Loss Landscapes}
To investigate the geometric influence of \met{} under well-behaved conditions, we visualize the local loss surface around the converged weights using two orthonormal directions in parameter space. The experiment is conducted under a stable training setup—DenseNet201 on Imagenette with a low learning rate ($1\mathrm{e}{-5}$).

\begin{figure}[h!]
  \centering
  \includegraphics[width=\linewidth]{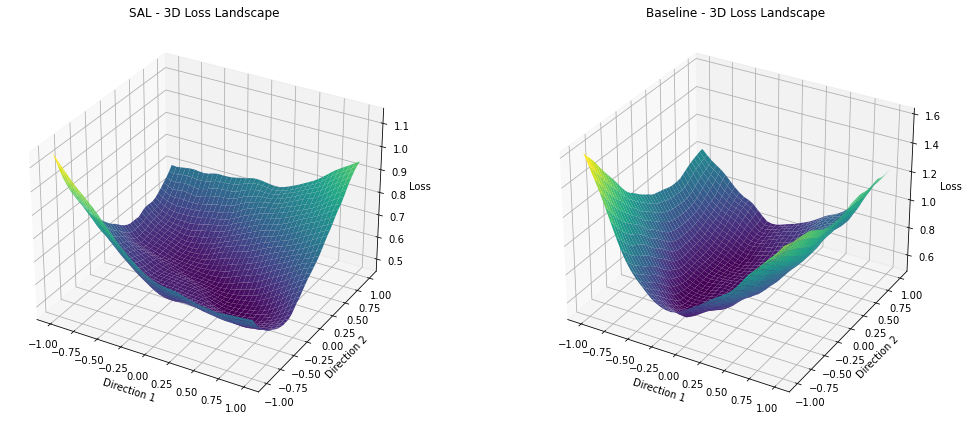}
  \caption{3D loss landscape under stable training. \met{} converges to a slightly wider and flatter region than the baseline, illustrating its ability to gently modulate training geometry even when stability is not a concern.}
  \label{fig:3d_loss_landscape}
\end{figure}

As shown in~\autoref{fig:3d_loss_landscape}, both the baseline and \met{} models converge to relatively smooth minima, which is expected given the stable setting. However, the region reached by \met{} appears marginally wider and flatter, with less pronounced curvature. This suggests that even under favorable conditions, stress-aware interventions of \met{} may encourage broader basins and milder curvature, contributing to improved generalization.

While the differences are not dramatic—as training does not stagnate severely in this regime—these results reinforce that \met{} does not disrupt learning unnecessarily. Instead, it adaptively promotes curvature regulation, subtly improving the geometry of convergence even when conditions are already stable.

\subsection{Evaluation Across Architectures and Efficiency}
\label{sec:Evaluation Across Architectures and Efficiency}
To assess the generality and practicality of \met{}, we benchmark it across a diverse set of architectures on Imagenette (with $64 \times 64$ resolution and a learning rate of $1\mathrm{e}{-5}$), as summarized in~\autoref{tab:arch_comparison}. This includes deep networks such as DenseNet201 and ResNet50, mid-sized architectures like DenseNet121 and EfficientNetB2, and lightweight mobile models such as MobileNetV2.

Across all tested models, \met{} consistently improves Top-1 accuracy, indicating robust generalization. DenseNet201, for instance, achieves an accuracy gain of 1.45 percentage points while reducing training time by nearly 40 percent, with no additional memory overhead. This highlights \met{}'s ability to enhance both efficiency and convergence quality in high-capacity models.

DenseNet121 achieves the highest Top-1 accuracy (88.56 percent) under \met{}, but with increased memory usage (from 4383 MB to 4690 MB) and a longer training duration (from 790 to 1372 seconds). This suggests that for certain mid-sized architectures, the benefits of stress-aware learning may require additional computational cost, yet remain feasible in practice.

In resource-constrained settings, \met{} shows clear advantages. For MobileNetV2, the method improves accuracy while significantly reducing memory usage (from 4375 MB to 1611 MB). Although training time increases, the final model remains lightweight and efficient. Similarly, EfficientNetB2 shows improved accuracy and reduced inference time under \met{}, with a substantial memory footprint reduction (from 1896 MB to 589 MB). These results suggest strong compatibility with mobile and edge computing requirements.

For ResNet50, \met{} provides a moderate accuracy gain with tolerable increases in training time and memory usage, demonstrating that the method scales reliably with standard residual networks.

In summary, \met{} delivers consistent performance improvements across a wide range of architectures, from deep and expressive networks to lightweight backbones. These gains are achieved without excessive computational burden, underscoring \met{}'s utility as a broadly applicable training enhancement for diverse real-world scenarios.

\begin{table*}[h!]
\centering
\caption{Comparison of Baseline (Adam) vs. \met{} across architectures on Imagenette ($64\times64$, 50 epochs).}
\resizebox{0.8\textwidth}{!}{%
\begin{tabular}{lccccccc}
\toprule
\textbf{Architecture} & \textbf{Model} & \textbf{Top-1 Acc (\%)} & \textbf{Top-5 Acc (\%)} & \textbf{Train Time (s)} & \textbf{Memory (MB)} & \textbf{Infer Time (ms)} \\
\midrule
\multirow{2}{*}{DenseNet201~\cite{huang2017densenet}} 
  & Baseline & 86.98 & 98.68 & 2593.29 & 1604.84 & 63.10 \\
  & \met{} & \textbf{88.43} & \textbf{98.73} & \textbf{1594.13} & 1604.84 & \textbf{52.15} \\
\midrule
\multirow{2}{*}{DenseNet121~\cite{huang2017densenet}} 
  & Baseline & 87.36 & \textbf{98.47} & \textbf{790.37} & \textbf{4383.05} & 44.65 \\
  & \met{} & \textbf{88.56} & 98.37 & 1372.66 & 4690.84 & \textbf{43.62} \\
\midrule
\multirow{2}{*}{MobileNetV2~\cite{sandler2018mobilenetv2}} 
  & Baseline & 82.96 & \textbf{98.06} & \textbf{382.64} & 4375.52 & \textbf{36.58} \\
  & \met{} & \textbf{83.46} & 97.89 & 657.95 & \textbf{1611.70} & 39.76 \\
\midrule
\multirow{2}{*}{ResNet50~\cite{he2016deep}} 
  & Baseline & 81.43 & 97.30 & \textbf{605.05} & \textbf{1961.63} & \textbf{40.55} \\
  & \met{} & \textbf{82.04} & \textbf{97.35} & 850.04 & 2075.77 & 44.20 \\
\midrule
\multirow{2}{*}{EfficientNetB2~\cite{tan2019efficientnet}} 
  & Baseline & 73.20 & 96.05 & \textbf{670.96} & 1896.67 & 48.59 \\
  & \met{} & \textbf{73.58} & \textbf{96.28} & 1103.69 & \textbf{589.64} & \textbf{44.18} \\
\bottomrule
\end{tabular}}
\label{tab:arch_comparison}
\end{table*}

\subsection{Evaluation Across Benchmarks}
\met{} demonstrates consistent improvements in generalization across a diverse range of vision benchmarks under a unified training setup using the Adam optimizer, a fixed learning rate of $1\mathrm{e}{-5}$, and $64 \times 64$ input resolution. As shown in~\autoref{tab:sal_top1_top5}, \met{} achieves higher Top-1 accuracy on all seven evaluated datasets and either improves or maintains Top-5 accuracy.

The gains vary in magnitude across benchmarks. On Imagenette, \met{} improves Top-1 accuracy from 86.98\% to 88.43\%, while also increasing Top-5 accuracy slightly. On the more challenging and visually diverse ImageNetWoof2 and Tiny-ImageNet datasets, the method still yields measurable improvements. For example, Tiny-ImageNet Top-1 accuracy rises from 58.64\% to 58.70\%, with a 0.58\% increase in Top-5 accuracy. Even modest gains in such tasks are meaningful, given the difficulty of achieving stable training.

On structured and higher-performing datasets such as EuroSAT and PACS, \met{} improves or preserves peak performance. For EuroSAT\_RGB, Top-1 accuracy increases by 0.5\%, while Top-5 accuracy remains unchanged at 99.93\%. On PACS, \met{} improves Top-1 accuracy from 96.20\% to 96.80\%, without impacting the perfect Top-5 performance. This confirms that \met{} does not over-regularize in high-accuracy regimes.

CIFAR-10 and Corel-1k, representing general-purpose and domain-generalization tasks respectively, also show positive trends. \met{} improves CIFAR-10 Top-1 accuracy by 0.35\% and Corel-1k by 1.5\%. The latter is particularly relevant given the known instability of training on small and noisy datasets, where \met{}'s stress-guided interventions appear to have greater impact.

Overall, these results confirm that \met{} is architecture- and dataset-agnostic, delivering measurable improvements across small-scale, imbalanced, and even high-performance benchmarks without additional tuning. Its stress-aware learning mechanism adapts seamlessly to data complexity, making it a robust option for general-purpose neural training.

\begin{table}[h!]
\centering
\caption{Top-1 and Top-5 accuracy comparison between Baseline and \met{} across multiple datasets. All models use the Adam optimizer, learning rate $1 \times 10^{-5}$, and $64 \times 64$ input resolution.}
\resizebox{0.48\textwidth}{!}{%
\begin{tabular}{lccc}
\toprule
\textbf{Benchmark} & \textbf{Model} & \textbf{Top-1 Acc (\%)} & \textbf{Top-5 Acc (\%)} \\
\midrule
\multirow{2}{*}{Imagenette~\cite{imagenette}} 
  & Baseline     & 86.98 & 98.68 \\
  & \met{}       & \textbf{88.43} & \textbf{98.73} \\
\midrule
\multirow{2}{*}{ImageWoof2~\cite{imagewoof}} 
  & Baseline     & 61.82 & 93.08 \\
  & \met{}       & \textbf{63.07} & \textbf{93.26} \\
\midrule
\multirow{2}{*}{Tiny-ImageNet~\cite{le2015tinyimagenet}} 
  & Baseline     & 58.64 & 77.78 \\
  & \met{}       & \textbf{58.70} & \textbf{78.36} \\
\midrule
\multirow{2}{*}{EuroSAT\_RGB~\cite{helber2019eurosat}} 
  & Baseline     & 97.17 & 99.93 \\
  & \met{}       & \textbf{97.67} & \textbf{99.93} \\
\midrule
\multirow{2}{*}{PACS~\cite{li2017deeper}} 
  & Baseline     & 96.20 & 100.00 \\
  & \met{}       & \textbf{96.80} & 100.00 \\
\midrule
\multirow{2}{*}{CIFAR-10~\cite{krizhevsky2009learning}} 
  & Baseline     & 90.24 & 99.41 \\
  & \met{}       & \textbf{90.59} & \textbf{99.42} \\
\midrule
\multirow{2}{*}{Corel-1k~\cite{corel1k}} 
  & Baseline     & 86.00 & 99.00 \\
  & \met{}       & \textbf{87.50} & \textbf{99.50} \\
\bottomrule
\end{tabular}
}
\label{tab:sal_top1_top5}
\end{table}

\subsection{Optimizer Adaptability}
\label{sec:optimizer_adaptability}
To assess the generalizability of \met{} across different optimization algorithms, we integrate it with Adam, Adamax, Nadam, and RMSProp using DenseNet201 on the Imagenette dataset ($64 \times 64$ input, 50 epochs). \autoref{tab:optimizer_adaptability} summarizes the results.

Across all optimizers, \met{} consistently improves both Top-1 and Top-5 accuracy compared to the baseline Adam configuration. RMSProp combined with SAL yields the highest Top-1 accuracy (89.17\%), indicating that SAL can leverage even momentum-based optimizers effectively. Adam with SAL offers the fastest training time, demonstrating efficient convergence under conservative optimization. Adamax and Nadam also benefit from SAL, with noticeable gains in performance despite their varying learning dynamics.

These results confirm that SAL operates in an optimizer-agnostic manner. It offers a flexible and robust enhancement mechanism regardless of the underlying optimizer, requiring no tuning of optimizer-specific parameters.

\begin{table*}[h!]
\centering
\caption{Performance of \met{} across various optimizers (DenseNet201 on Imagenette, $64\times64$ resolution, 50 epochs).}
\resizebox{0.7\textwidth}{!}{%
\begin{tabular}{lcccc}
\toprule
\textbf{Optimizer} & \textbf{Top-1 Acc (\%)} & \textbf{Top-5 Acc (\%)} & \textbf{Train Time (s)} & \textbf{Memory (MB)} \\
\midrule
Adam~\cite{kingma2014adam} (Baseline)       & 86.98 & 98.68 & 2593.29 & 1604.84 \\
\midrule
Adam~\cite{kingma2014adam} in \met{} & 88.43 & 98.73 & 1594.13 & 1604.84 \\
\midrule
Adamax~\cite{kingma2014adam} in \met{} & 87.59 & 98.68 & 2254.41 & 4649.41 \\
\midrule
Nadam~\cite{dozat2016incorporating} in \met{}  & 88.51 & 98.73 & 3128.37 & 959.20  \\
\midrule
RMSProp~\cite{tieleman2012lecture} in \met{}  & 89.17 & 98.75 & 2568.85 & 1439.95 \\
\bottomrule
\end{tabular}}
\label{tab:optimizer_adaptability}
\end{table*}

\section{Limitations and Future Work}

Within the scope of this work, \met{} demonstrates strong adaptability and effective training regulation via a global stress signal and targeted parameter interventions. While the current design performs well across diverse benchmarks and architectures in computer vision, several directions remain for future refinement.

One avenue is extending stress modeling to a layer-wise or module-specific level, enabling more localized control and adaptive plasticity throughout the network. Presently, plastic deformation is applied to the upper layers, balancing efficiency with effectiveness; however, dynamically selecting intervention layers based on stress patterns may enhance flexibility and generalization.

Additionally, the current empirical evaluation, while comprehensive across architectures and datasets, could benefit from further experimental depth. Future studies may include multiple runs with statistical significance testing and confidence intervals to solidify the observed gains. Conducting systematic ablation studies that isolate the individual effects of stress signal accumulation, Gaussian noise injection, and plastic deformation would help clarify the contribution of each component.

It is also essential to compare \met{} against alternative adaptive training paradigms with similar objectives, such as SAM~\cite{foret2020sharpness}, to better position its unique advantages and limitations. 

While this work focuses exclusively on vision tasks, the SAL framework is inherently domain-agnostic and could potentially benefit other learning paradigms. Future research may explore its application in Reinforcement Learning (RL) \cite{mahdavi2022rl}, Natural Language Processing (NLP), and Large Language Models (LLMs), where dynamic training regulation and stress-aware interventions may help mitigate issues such as vanishing gradients, reward sparsity, or catastrophic forgetting. Investigating the behavior of accumulated stress under temporal or sequential input structures could yield valuable extensions of SAL in these domains.

Moreover, integrating SAL with advanced generative architectures such as Diffusion Models \cite{farshad2023scenegenie}, and scaling it to large-scale datasets like ImageNet or multi-modal settings could provide further evidence of its generalizability and robustness.

\section{Conclusion}
We proposed \met{}, a stress-regulated training framework inspired by plastic deformation. It introduces adaptive interventions—mild noise or stronger parameter shifts—based on accumulated training stress, promoting robustness and convergence. Experiments across various models, datasets, optimizers, and resolutions show that \met{} improves generalization and stability, particularly under unstable conditions. As a lightweight, feedback-driven method, SAL offers a promising direction for resilient deep learning systems.

\small
\bibliographystyle{unsrt}
\bibliography{references}

\appendix
\label{sec:appendix}
\section{Programming Environment}
All experiments were implemented in Python using TensorFlow 2.15 and Keras APIs. Training and evaluation were performed on a local workstation equipped with an AMD Ryzen 7 5800H CPU, 16GB RAM, and an NVIDIA GeForce RTX 3060 GPU with 6GB VRAM. The system ran Windows 11 with CUDA 11.8 and cuDNN 8.6 properly configured. All numerical computations used 32-bit floating-point precision. Memory consumption was monitored using the \texttt{psutil} package, and data pipelines were parallelized via \texttt{tf.data.AUTOTUNE} for optimal throughput.

\section{Extended Results}
\subsection{Stress Behavior Distribution}
\begin{figure}[h!]
  \centering
  \includegraphics[width=0.75\linewidth]{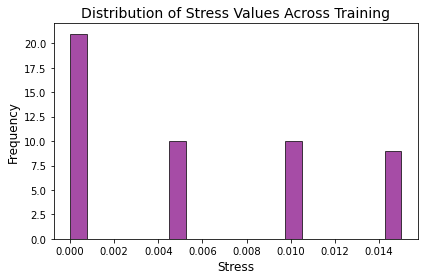}
  \caption{Distribution of accumulated stress values across training epochs. Most epochs lie in low-to-moderate stress zones, reflecting stable training. Peaks near the critical threshold denote rare but decisive plastic deformation events.}
  \label{fig:stress_distribution}
\end{figure}

~\autoref{fig:stress_distribution} presents the empirical distribution of the accumulated stress signal $S_g$ throughout training. The histogram reveals a skewed distribution, with the majority of epochs residing in the low to moderate stress regime. This reflects stable learning dynamics, where minor perturbations suffice to sustain generalization.

A smaller proportion of epochs exhibit elevated stress values approaching the critical yield threshold $S_{\text{yield}}$. These high-stress phases, though infrequent, are pivotal: they correspond to plastic deformation events that strategically redirect the optimization trajectory away from sharp or stagnant minima.

The scarcity of high-stress intervals and the dominant presence of low-stress regions support a key design principle of \met{}: interventions should be selective and proportional, not uniformly applied. This confirms that \met{} avoids unnecessary disruption and adapts its behavior based on the actual difficulty encountered during training. Overall, the distribution validates the system's ability to operate in a regulated, context-aware manner.

\subsection{Accuracy Gain Over Baseline}
\begin{figure}[h!]
  \centering
  \includegraphics[width=0.75\linewidth]{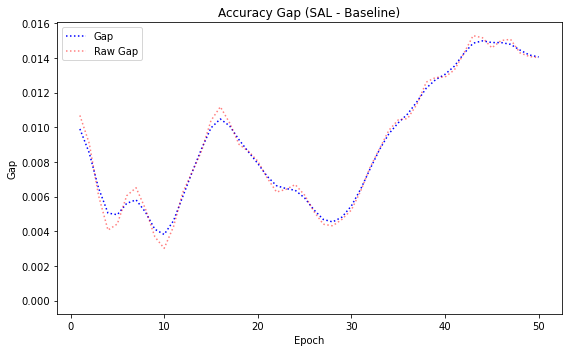}
  \caption{Accuracy gap between \met{} and baseline optimizer across epochs. The consistently positive and increasing gap supports the cumulative benefits of stress-informed adaptation.}
  \label{fig:accuracy_gap_plot}
\end{figure}

~\autoref{fig:accuracy_gap_plot} depicts the evolution of the accuracy gap between \met{} and the baseline optimizer throughout training. The gap remains consistently positive, confirming that \met{} outperforms its non-adaptive counterpart at nearly every epoch. Notably, the gap widens after epoch 30, indicating that the advantages of stress-aware regulation compound over time. The smoothed trend further emphasizes this systematic improvement. These results demonstrate that \met{} not only achieves superior final performance but also maintains a more stable and robust learning trajectory throughout training.

\subsection{Trajectory Analysis in Weight Space}
\begin{figure}[h!]
  \centering
  \includegraphics[width=0.75\linewidth]{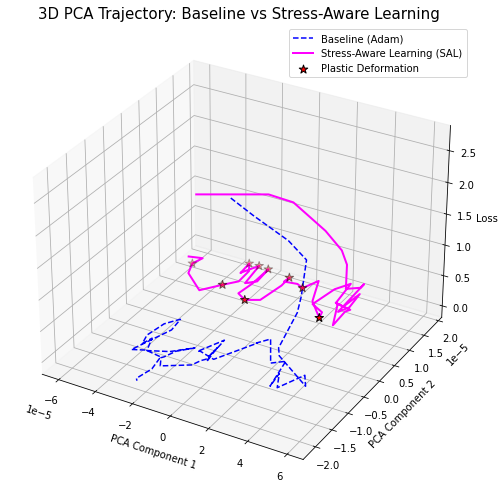}
  \caption{PCA-reduced 3D trajectory of weight updates across training. The \met{} trajectory (magenta) explores broader, curved regions compared to the baseline (blue). Inflection points in the path coincide with stress-triggered interventions, showing how \met{} dynamically reshapes the optimization route.}
  \label{fig:3d_trajectory}
\end{figure}

To understand how \met{} influences the optimizer's trajectory, we project the high-dimensional weight updates into a 3D space using Principal Component Analysis (PCA). As shown in ~\autoref{fig:3d_trajectory}, the baseline optimizer follows a relatively linear and confined path, indicative of limited exploration and a tendency to converge into local minima.

In contrast, the trajectory induced by \met{} is more expansive and nonlinear, traversing broader regions of the parameter space. Notably, the trajectory exhibits pronounced deflections at epochs corresponding to elevated stress levels—precisely when plastic interventions are applied. These deviations suggest that stress-triggered perturbations not only improve local escape but also reorient the optimizer toward more favorable convergence basins.

This behavior aligns with the design objectives of \met{}: by integrating stress-aware deformation, the training path becomes more resilient to premature convergence and better aligned with generalizable solutions. The spatial divergence between trajectories further supports that \met{} enhances both exploration and robustness.

\subsection{Behavior Across Image Resolutions}
To evaluate the resolution sensitivity of \met{}, we train DenseNet201 models under identical configurations while varying the input resolution on the Imagenette dataset. As reported in~\autoref{tab:image_resolution}, \met{} maintains strong generalization performance across all tested sizes, from coarse $32 \times 32$ inputs to high-resolution $128 \times 128$ images.

The results demonstrate a clear trend: as resolution increases, both Top-1 and Top-5 accuracy improve, reaching 97.17\% and 99.82\% respectively at $128 \times 128$. Notably, even at lower resolutions (e.g., $64 \times 64$ and $32 \times 32$), where training typically becomes more volatile due to reduced spatial detail, \met{} preserves stable learning dynamics and avoids performance collapse. Specifically, at $32 \times 32$, it still achieves 70.29\% Top-1 accuracy, outperforming common baselines reported at similar resolutions.

This analysis underscores a key property of \met{}: its ability to adaptively stabilize training regardless of resolution-induced difficulty. By injecting stress-aware perturbations during low-information regimes, \met{} recovers meaningful convergence where conventional optimizers might stall. These findings further support the general-purpose applicability of Stress-Aware Learning to real-world scenarios with resource or resolution constraints.

\begin{table}[h!]
\centering
\caption{Performance of \met{} across different input resolutions on Imagenette.}
\resizebox{0.45\textwidth}{!}{%
\begin{tabular}{ccc}
\toprule
\textbf{Image Resolution} & \textbf{Top-1 Acc (\%)} & \textbf{Top-5 Acc (\%)} \\
\midrule
$128 \times 128$ & 97.17 & 99.82 \\
$96 \times 96$   & 94.70 & 99.69 \\
$64 \times 64$   & 88.43 & 98.73 \\
$32 \times 32$   & 70.29 & 95.39 \\
\bottomrule
\end{tabular}}
\label{tab:image_resolution}
\end{table}

\subsection{Class-Level Performance}
\label{sec:classification_stable}
To further investigate the impact of \met{} under the stable training condition (\autoref{sec:Stable Training Conditions}), we report detailed classification metrics on Imagenette ($64\times64$ resolution) in~\autoref{tab:classification_comparison}. Results reveal that \met{} improves performance across most categories, particularly in F1-score, which reflects a balanced improvement in both precision and recall.

Notably, \met{} yields consistent gains in 9 out of 10 classes, with particularly strong improvements in class 2 (F1-score: 88.64\% $\rightarrow$ 91.57\%) and class 3 (F1-score: 87.64\% $\rightarrow$ 89.26\%). These improvements are not marginal; they highlight the model’s ability to sustain learning progress during stagnation phases. Precision and recall also show substantial gains, indicating that \met{} reduces false positives and false negatives simultaneously. The macro-averaged metrics (Macro Avg) show a uniform performance boost across all classes, with precision, recall, and F1-score increasing from 86.96\% to 88.43\%. Weighted averages (Weighted Avg) follow the same trend, confirming that the improvements are not limited to underrepresented classes but extend to the overall class distribution.

These gains are achieved without altering the model architecture, training data, or increasing computational complexity—demonstrating that stress-aware interventions alone can lead to more balanced and generalizable classification performance.

\begin{table*}[h!]
\centering
\caption{Classification comparison of Baseline vs. \met{} on Imagenette ($64\times64$ resolution).}
\resizebox{1\textwidth}{!}{%
\begin{tabular}{lcccccccc}
\toprule
\multirow{2}{*}{\textbf{Class}} & \multicolumn{4}{c}{\textbf{Baseline}} & \multicolumn{4}{c}{\textbf{\met{}}} \\
\cmidrule(lr){2-5} \cmidrule(lr){6-9}
 & Precision (\%) & Recall (\%) & F1-score (\%) & Support & Precision (\%) & Recall (\%) & F1-score (\%) $\uparrow$ & Support \\
\midrule
Tench & 93.09 & 90.44 & 91.74 & 387 & 89.66 & 94.06 & 91.80 & 387 \\
English springer & 88.41 & 88.86 & 88.64 & 395 & 93.88 & 89.37 & 91.57 & 395 \\
Cassette player & 85.98 & 89.36 & 87.64 & 357 & 87.80 & 90.76 & 89.26 & 357 \\
Chain saw & 72.32 & 78.50 & 75.28 & 386 & 77.50 & 80.31 & 78.88 & 386 \\
Church & 90.12 & 91.44 & 90.78 & 409 & 88.10 & 94.13 & 91.02 & 409 \\
French horn & 88.62 & 85.03 & 86.79 & 394 & 90.48 & 86.80 & 88.60 & 394 \\
Garbage truck & 83.33 & 87.40 & 85.32 & 389 & 85.47 & 90.75 & 88.03 & 389 \\
Gas pump & 83.85 & 76.85 & 80.20 & 419 & 84.25 & 76.61 & 80.25 & 419 \\
Golf ball & 94.67 & 93.48 & 94.07 & 399 & 95.38 & 93.23 & 94.30 & 399 \\
Parachute & 90.34 & 88.72 & 89.52 & 390 & 92.53 & 88.97 & 90.72 & 390 \\
\midrule
\textbf{Top-1 Acc} & \multicolumn{4}{c}{86.96} & \multicolumn{4}{c}{\textbf{88.43}} \\
\textbf{Macro Avg} & 87.07 & 87.01 & 87.00 & 3925 & \textbf{88.51} & \textbf{88.50} & \textbf{88.44} & 3925 \\
\textbf{Weighted Avg} & 87.11 & 86.96 & 86.99 & 3925 & \textbf{88.51} & \textbf{88.43} & \textbf{88.41} & 3925 \\
\bottomrule
\end{tabular}}
\label{tab:classification_comparison}
\end{table*}

\end{document}